\documentclass[review,11pt]{ReportTemplate}
\usepackage{bm}
\usepackage{graphicx}
\usepackage{amsmath,mathrsfs,amssymb}

\newtheorem{definition}{Definition}
\DeclareMathOperator*{\argmax}{arg\,max}

\begin{document}
\begin{frontmatter}
\title{Tunneling Neural Perception and Logic Reasoning through Abductive Learning}

\author{Wang-Zhou Dai$^\dag$}
\author{Qiu-Ling Xu$^\dag$}
\author{Yang Yu$^\dag$}
\author{Zhi-Hua Zhou\corref{cor1}}
\address{National Key Laboratory for Novel Software Technology\\
Nanjing University, Nanjing 210093, China} \cortext[cor1]{\small Corresponding author. 
Email: zhouzh@nju.edu.cn\\
\hspace*{1.22em}$^\dag$These authors contributed equally to this work}

\begin{abstract}
  Perception and reasoning are basic human abilities that are seamlessly connected as part of human intelligence. However, in current machine learning systems, the perception and reasoning modules are incompatible. Tasks requiring joint perception and reasoning ability are difficult to accomplish autonomously and still demand human intervention. Inspired by the way language experts decoded Mayan scripts by joining two abilities in an abductive manner, this paper proposes the abductive learning framework. The framework learns perception and reasoning simultaneously with the help of a trial-and-error abductive process. We present the Neural-Logical Machine as an implementation of this novel learning framework. We demonstrate that---using human-like abductive learning---the machine learns from a small set of simple hand-written equations and then generalises well to complex equations, a feat that is beyond the capability of state-of-the-art neural network models. The abductive learning framework explores a new direction for approaching human-level learning ability.
\end{abstract}

\begin{keyword}
Machine Learning \sep logic \sep neural network \sep perception \sep
abduction \sep reasoning
\end{keyword}
\end{frontmatter}

Mayan scripts were a complete mystery to modern humanity until its numerical systems and calendars were first successfully deciphered in the late 19th century. As described by historians, the number recognition was derived from a handful of images that show mathematical regularity~\cite{mayabook}. The decipherment was not trivial because the Mayan numerical system is vigesimal (base twenty), totally different from the decimal system currently in common use. The successful deciphering of Mayan numbers reflects two remarkable human intelligence capabilities: 1) visually perceiving individual characters from images and 2) reasoning symbolically based on mathematical background knowledge during perception. These two abilities function at the same time and affect each other. Moreover, the two abilities are often joined subconsciously by humans, which is key in many real-life learning problems.

Modern artificial intelligence (AI) systems exhibit both these abilities---but only in isolation. Deep neural networks have achieved extraordinary performance levels in recognizing human faces~\cite{Taigman_2014_CVPR}, objects~\cite{Krizhevsky2012ImageNet,NIPS2013_5207}, and speech~\cite{Hinton2012Deep}; meanwhile, logic-based AI systems have achieved human-level abilities in proving mathematical theorems~\cite{NewellS56,Chang:1997:SLM} and in performing inductive reasoning concerning relations~\cite{mugg:ilp}. However, recognition systems can hardly exploit complex domain knowledge in symbolic forms, perceived information is difficult to include in reasoning systems, and a reasoning system usually requires semantic-level knowledge, which involves human input~\cite{russell:unify}. Even in recent neural network models with enhanced memories~\cite{Graves2016Hybrid}, the ability to focus on relations~\cite{Santoro17relation}, and differentiable knowledge representations~\cite{hinton:learndistrep,hu2016harnessing,GauntBKT17,BosnjakRNR17}, full logical reasoning ability is still missing---as an example, consider the difficulties of understanding natural language~\cite{Jia2017EMNLP}. To glue together perception and reasoning, it is crucial to answer the question: How should perception and reasoning affect one another in a single system?


\subsection*{Mayan Hieroglyph Decipherment}
\begin{figure}[!ht]
    \centering
    \includegraphics[width=1\linewidth]{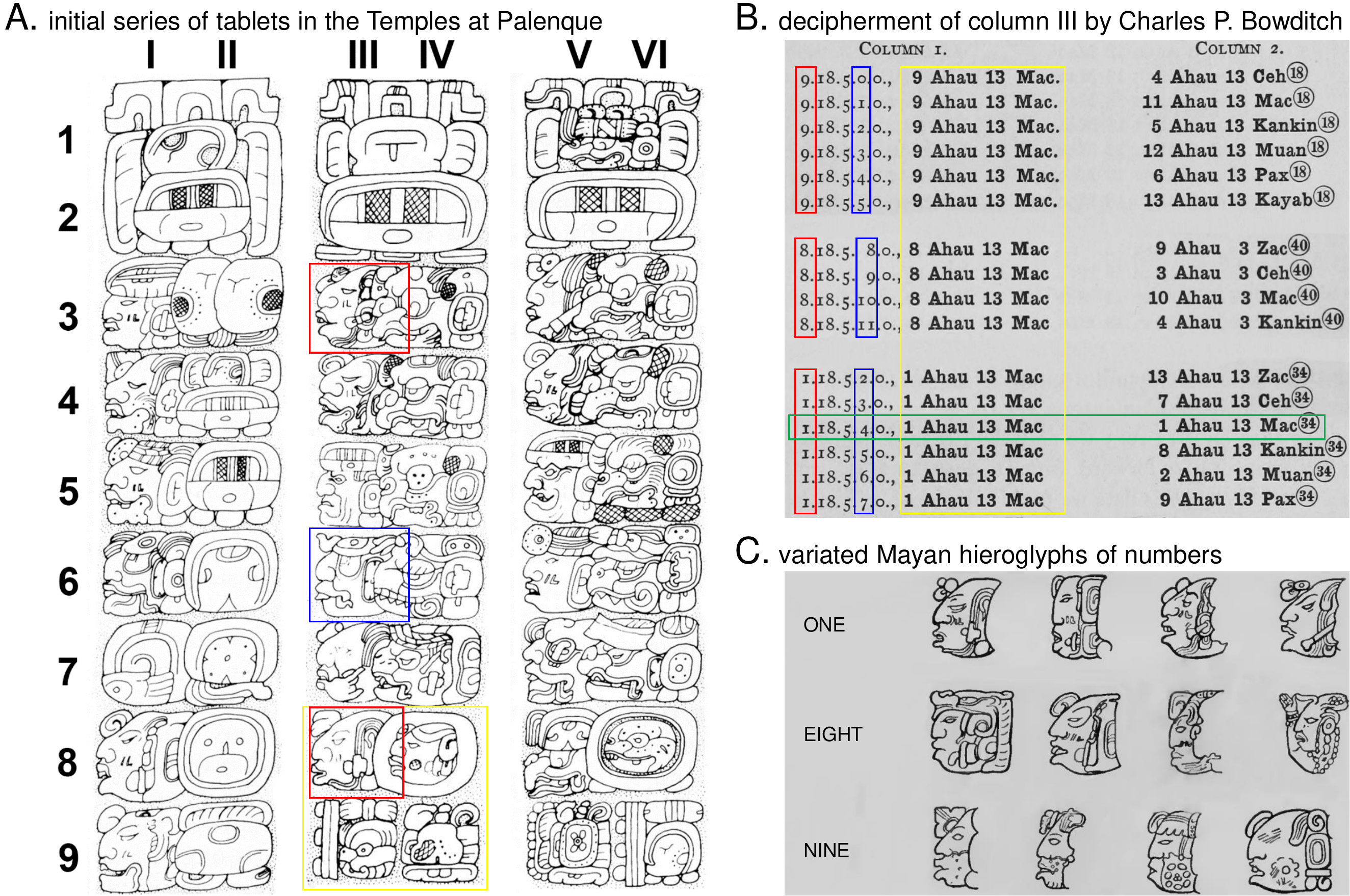}
    \caption{{\bf Illustration of Mayan hieroglyph decipherment.} 
    {\bf (A)} In rows 3---9, the odd columns represent numbers and the even columns represent calendar time units. Column II shows the standard representations for the units; columns IV and VI are identical unit representations but in different drawings. Rows 1---2 are initials, rows 3---6 represent time spans, and rows 8---9 are dates computed from the time spans. The hieroglyphs marked by boxes correspond to the numbers and units in the same colored boxes in subfigure (B). {\bf (B)} Column 1 lists possible interpretations from Column III of subfigure (A), and Column 2 lists the results calculated from the numbers in Column 1. Bowditch first identified the hieroglyphs at III4, III5, III7 and III9 in (A) and then confirmed that III3 and III9 represent the same numbers. Initially, he abduced III3 as 9 based on his past experience with Mayan calendars; however, that was impossible because the calculated results were inconsistent with the dates. Then, he tried substituting those positions with numbers that have similar hieroglyphs. Finally, he confirmed that the interpretation ``1.18.5.4.0 1 Ahau 13 Mac'' (in the green box) should be correct. It is very unusual for “1” to be attached to the unit in row 3, but its presence there is confirmed by its consistency with subsequent passages in the same tablet. {\bf (C)} The highly varied character representations and unusual calendar system cause the decipherment of Mayan hieroglyphs to require both sensitive vision and a logical mind. Credits: subfigure (A) is reproduced from~\cite{mayatablets}; subfigures (B) and (C) are reproduced from~\cite{bowitchbook}.}
    \vspace{5em}
    \label{fig:puzzle}
\end{figure}
In a quest for the answer, we return to the process by which Charles P. Bowditch deciphered Mayan numbers, which were inscribed as the heads of gods (now known as \emph{head variants} of numbers)~\cite{bowitchbook}. Figure~\ref{fig:puzzle} illustrates this process. Figure~\ref{fig:puzzle}(A) displays parts of three tablets discovered at Palenque. The first tablet (columns I---II) uses standard hieroglyphs to represent Mayan time units, e.g., ``Tun'' (360 days) at II5 and as parts of the initials in row 2 on all three tablets. Columns IV and VI draw the units in a totally different way, but Bowditch conjectured that they were identical to column II based on their positions. Moreover, although the Mayan numeral system is vigesimal, the unit ``Tun'' is just 18 times its predecessor, ``Winal'' (row 6 in all the even columns), making the decipherment even more difficult. Bowditch verified this through calculations and by evaluating the consistency of the relationships in these tablets. Then, he started to decipher the numbers in column III. As illustrated in Figure~\ref{fig:puzzle}(B), by mapping the hieroglyphs to different numbers and checking whether these numbers were consistent under the mathematical rules, Bowditch finally decoded the numbers and proved their correctness~\cite{bowitchbook}.

Bowditch's decipherment of the Mayan hieroglyphs explicitly illustrates the key aspect of joint visual perception and logical reasoning: in this case, a tunnel between perception and reasoning was established through a  \emph{trial-and-error} process of the hieroglyphic interpretations as shown in Column 1 in Figure~\ref{fig:puzzle}(B). The \emph{trial} step perceives, interprets the picture, and passes the interpreted symbols for consistency checking, while the \emph{error} step evaluates the consistency, uses reasoning to find errors in the interpretation, and provides error feedback to correct the perception.

This problem-solving process was called ``abduction'' by Charles S. Peirce~\cite{peirce:abduction} and termed ``retro-production'' by Herbert A. Simon~\cite{simon:retro}; it refers to the process of selectively inferring certain facts and hypotheses that explain phenomena and observations based on background knowledge~\cite{Magnani:2009:abd,KakasF09Abd}. In Bowditch’s Mayan number decipherment, the background knowledge involved arithmetic and some basic facts about Mayan calendars; the hypotheses involved a recognition model for mapping hieroglyphs to meaningful symbols and a more complete understanding of the Mayan calendar system. Finally, the validity of the hypotheses was ensured by trial-and-error searches and consistency checks.

\subsection*{Overview of the Abductive Learning Framework}
Inspired by the human abductive problem-solving process, we propose the \emph{Abductive Learning} framework to enable knowledge-involved joint perception and reasoning capability in machine learning.

Generally, machine learning is a process that involves searching for an optimal model within a large hypothesis space. Constraints are used to reduce the search space. Most of the machine learning algorithms exploit constraints expressed explicitly through mathematical formulations. However, as was the case with the domain knowledge used in Mayan language decipherment, many complex constraints in real-world tasks take the form of symbolic rules. Moreover, such symbolic knowledge can be incomplete or even inaccurate. \emph{Abductive Learning} uses \emph{logical abduction}~\cite{KakasF09Abd} to handle the imperfect symbolic inference problem. Given domain knowledge written as first-order logical rules, \emph{Abductive Learning} can abduce multiple hypotheses as possible explanations to observed facts, just as Bowditch made guesses about the unknown hieroglyphs based on his knowledge of arithmetic and Mayan language during his ``trial'' steps.

To exploit domain knowledge written as first-order logical rules, traditional logic-based AI uses the rules to make logical inferences based on input logical groundings, which are logical facts about the relations between objects in the domain. This, in fact, implicitly assumes the absolute existence of both the objects and the relations. However, as Stuart Russell commented, ``real objects seldom wear unique identifiers or pre-announce their existence like the cast of a play''~\cite{russell:unify}. Therefore, \emph{abductive learning} adopts neural perception to automatically abstract symbols from data; then, the logic abduction is applied to the generalized results of neural perception.

The key to \emph{abductive learning} is to discover how logical abduction and neural perception can be trained together. More concretely, when a differentiable neural perception module is coupled to a non-differentiable logical abduction module, learning system optimization becomes extremely difficult: the traditional gradient-based methods are inapplicable. In analogy to Bowditch’s decipherment, \emph{abductive learning} combines the two functionalities using a heuristic trial-and-error search approach.

Logical abduction, as a discrete reasoning system, can easily address a set of symbolic inputs. The neural layers involved in perception should output symbols that make the symbolic hypotheses consistent with each other. When the hypotheses are inconsistent, the logical abduction module finds incorrect output from the neural perception module and corrects it. This process is exactly the trial-and-error process that Bowditch followed in Figure~\ref{fig:puzzle}(B). The corrections function as the supervised signals to train the neural perception.

\begin{figure}[h!]
    \centering
    \includegraphics[width=\linewidth]{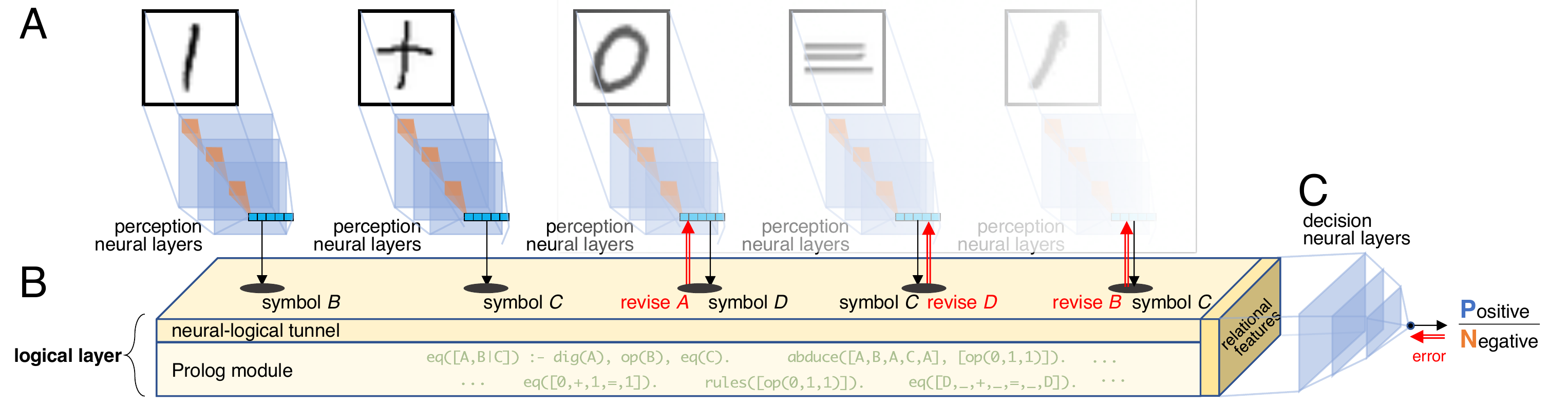}
    \caption{{\bf The architecture of a neural-logical machine}. {\bf (A)} Perception neural layers (such as convolutional layers) accomplish the perception task. {\bf (B)} The perception results are the input for the logical layer, which consists of the neural-logical tunnel, Prolog module, and relational features. The Prolog module checks the input consistency and produces relational features; the neural logical tunnel corrects the perception output based on consistency with the hypotheses; and the relational features expose the logical process outcomes. {\bf (C)} The decision neural layers transform the relational features into the final output.}
    \label{fig:architecture}
\end{figure}

To verify the effectiveness of \emph{abductive learning}, we implemented a Neural Logical Machine (NLM) as a demonstration of the \emph{abductive learning} framework. The architecture of an NLM for classifying handwritten equations is shown in Figure~\ref{fig:architecture}. The equations consist of sequential pictures of characters, as in the examples shown in Figure~\ref{fig:data}. An equation is associated with a label (positive or negative) that indicates whether the equation is correct or incorrect. A machine is tasked with learning from a training set of labeled equations, and the trained model is expected to predict future equations correctly. This task simulates Mayan hieroglyph decipherment: the machine does not know the meaning of the character pictures or the calculation rules in advance. Thus, this task demands the same ability as a human jointly utilizing perceptual and reasoning abilities.

Deep neural networks have been demonstrated to have incomparable perception performance on images~\cite{Krizhevsky2012ImageNet}. Our implementation of NLM employs a convolutional neural network (CNN)~\cite{lecun1998gradient} as the perception neural layers. The CNN takes image pixels as input and is expected to output the symbols in the image. The symbol output forms the input to the logical layer. To process the symbols logically and efficiently, the core of the logical layer is a Prolog module. Prolog is a powerful general-purpose logic programming language rooted in first-order logic. A common limitation of logic-based learning is its lack of flexibility when dealing with the uncertainty (such as noise and system errors) that exists in the real-world. Thus, we do not require the logical layer to output the final prediction directly. Instead, the logical layer outputs the values of some relational features that reflect the deductions made inside the Prolog module. Finally, the relational feature values are fed into the decision neural layers, which are implemented as a fully-connected multilayer feedforward neural network. The decision neural layers handle the uncertainty that exists between the logical outcomes and the labels.

The heuristic trial-and-error search is implemented using derivative-free optimization~\cite{Yu2016Derivative} in the neural-logical tunnel. Although the logical layer can find inconsistencies between the logic rules and the perceived symbols, it cannot find the positions of the incorrectly perceived symbols. NLM employs a derivative-free optimization method~\cite{Yu2016Derivative} to intelligently guess the positions at which the symbols appear incorrectly. For each guess, the Prolog module runs the abductive logical programming (ALP)~\cite{KakasKT92ALP} process that abduces whether the correct symbols appear at the indicated positions, making the logical hypotheses more consistent. We further accelerate the NLM by feeding it only a sample of the available training data during each training iteration. From a dataset sample, we can obtain only locally consistent hypotheses. Finally, the NLM transforms the locally consistent hypotheses into relational features using the propositionalization technique~\cite{kramer:propositionalisation}.

As an analogy to human abductive problem-solving, NLM works as follows. Before training, domain knowledge---written as a first-order logic program---is provided to the Prolog module. In our implementation, this background knowledge involves only the logic structure rules, as shown in Figure~\ref{fig:BK}. After training starts, a sample of the training data will be interpreted to candidate primitive symbols pre-defined in the neural-logical tunnel. Because the perception neural layer is initially a random network, the interpreted symbols are typically wrong and form inconsistent hypotheses. The logical layer starts to revise the interpreted symbols and search for the most consistent logical hypothesis in the training data sample. The hypotheses are stored as relational features in the logical layer, while the symbol revisions are used to train the perception neural layer in a straightforward supervised manner. When the training of these two subparts is complete (e.g., the perception layer converges or reaches an iteration limit), all the training examples are processed again by the NLM to obtain their feature vectors with regard to the abduced relational features. Finally, the decision neural layer is trained with these feature vectors from the whole dataset. The decision neural layer learning process will automatically filter ill-performing perception neural layer, hypotheses, and relational features. Moreover, due to the high complexity of symbolic abduction, we adopt the curriculum learning paradigm for training NLM (i.e., it begins learning from easier examples, and the difficulty of the learning tasks is gradually increased~\cite{Bengio2009Curriculum}.

\subsection*{Preliminaries}
\textbf{Logic Programming} is a type of programming paradigm that is largely based on formal logic. It is designed for symbolic computation and is especially well suited for solving problems that involve objects and the relations between them~\cite{kowalski74}.

One of the most widely used logic programming language is Prolog~\cite{Bratko:1990:Prolog}, which is designed based on first-order logic. A Prolog program consists of a set of logical facts and rules. For example, the fact that ``Adam is the father of Bob'' can be written in Prolog as:
\begin{quote}
    \vspace{-2em}
    \texttt{father(adam, bob).}
    \vspace{-2em}
\end{quote}
Here the \texttt{father} is the name of a property, called as \emph{predicate}; \texttt{adam} and \texttt{bob} are its arguments. A Rule stating that ``if father of \texttt{B} is \texttt{A}, then \texttt{A} is also a parent of \texttt{B}'' can be written as:
\begin{quote}
    \vspace{-2em}
    \texttt{parent(A, B) :- father(A, B).}
    \vspace{-2em}
\end{quote}
Here the ``\texttt{:-}'' denotes logical implication; \texttt{A} and \texttt{B} are logical variables.

By using Selective Linear Definite (SLD) clause resolution~\cite{kowalskiK71}, Prolog can perform first-order logical inferences. For example, given the above facts and rules as a logical program, the following question could be asked of the Prolog system:
\begin{quote}
    \vspace{-2em}
    \texttt{?- parent(X, bob).}
    \vspace{-2em}
\end{quote}
Having access to the previously asserted fact, Prolog will answer:
\begin{quote}
    \vspace{-2em}
    \texttt{X = adam.}
    \vspace{-2em}
\end{quote}
However, if the query is:
\begin{quote}
    \vspace{-2em}
    \texttt{?- parent(eve, bob).}
    \vspace{-2em}
\end{quote}
Prolog will answer
\begin{quote}
    \vspace{-2em}
    \textbf{false.}
    \vspace{-2em}
\end{quote}
because the previous program does not specify any relation between \texttt{eve} and \texttt{bob}.

Owing to its comprehensibility and the power of performing first-order logical inferences, logic programming is widely used in symbolic AI systems such as expert systems~\cite{Chang:1997:SLM}, inductive logic programming~\cite{mugg:ilp}, abductive logic programming~\cite{KakasKT92ALP}, and so on.

\noindent\textbf{Derivative-free optimization}, as a counterpart of gradient-based optimization methods, solves optimization tasks without requiring the derivative information of the optimization function. Instead, it uses sampling methods to draw samples from the solution space and learns a potential region from which further samples will be drawn. Recent studies have shown that derivative-free optimization algorithms can solve a range of sophisticated optimization functions at a guaranteed level~\cite{MunosFTML2014,Wang16,Yu2016Derivative}. This work thus employs a state-of-the-art derivative-free optimization approach to solve the raised non-differentiable functions.

\subsection*{Problem Setting}
The input of abductive learning consists of a set of labeled training data $X=\{\langle x_1,y_1\rangle,\ldots,\langle x_n,\\y_n\rangle\}$ about target concept $C$ and domain knowledge $T$, where $x_i\in \mathbb{R}^m$ is a raw feature space, $y_i\in\{0,1\}$ is the label for $x_i$ on target concept $C$, and $T$ is a logical theory expressed by a set of first-order logical formulas. 

In contrast to ordinary statistical machine learning problems, the target concept $C$ describes a certain relationship between a set of primitive concepts $P=\{p_1,\ldots,p_r\}$ in the domain; thus, it can hardly be directly induced from the raw feature space $\mathbb{R}^m$ using statistical models. Therefore, learning the target concept requires symbolic reasoning based on the set of primitive concepts $P$ in the logical theory $T$. However, although the primitive concepts $P$ are defined in $T$, the mapping of $p(x):\mathbb{R}^m\mapsto P$ from the raw feature space to the primitive concepts is unknown. Furthermore, the domain knowledge in $T$ is incomplete, i.e., it is some missing logic formulas $\Delta_C$ that describe the relations between primitive concepts $P$ that are required as complements to $T$ to define the target concept. The target output of abductive learning is to learn the mapping for $p(x)$ and the symbolic knowledge $\Delta_C$ simultaneously from the data $X$, where $p(x)$ and $\Delta_C$ are respectively called the \emph{perception model} and the \emph{reasoning model} in this paper.

For example, in the binary additive equation learning problem in Figure~\ref{fig:data}: $X$ consists of images of equations and their labels; $T$ contains basic arithmetic knowledge but no specific calculation rules for calculating ``addition''; the primitive concepts of $P$ are the digit and operator symbols $\{0,1,+,=\}$. The goal is to learn a perception model $p$ mapping images to the symbols, and a reasoning model $\Delta_C$ of addition rules for calculating ``$+$'' operations, such as arithmetic calculations or logical exclusive-or operations.

\subsubsection*{Neural-Logical Machine Implementation}

\begin{figure}[t]
    \centering
    \includegraphics[width=0.6\linewidth]{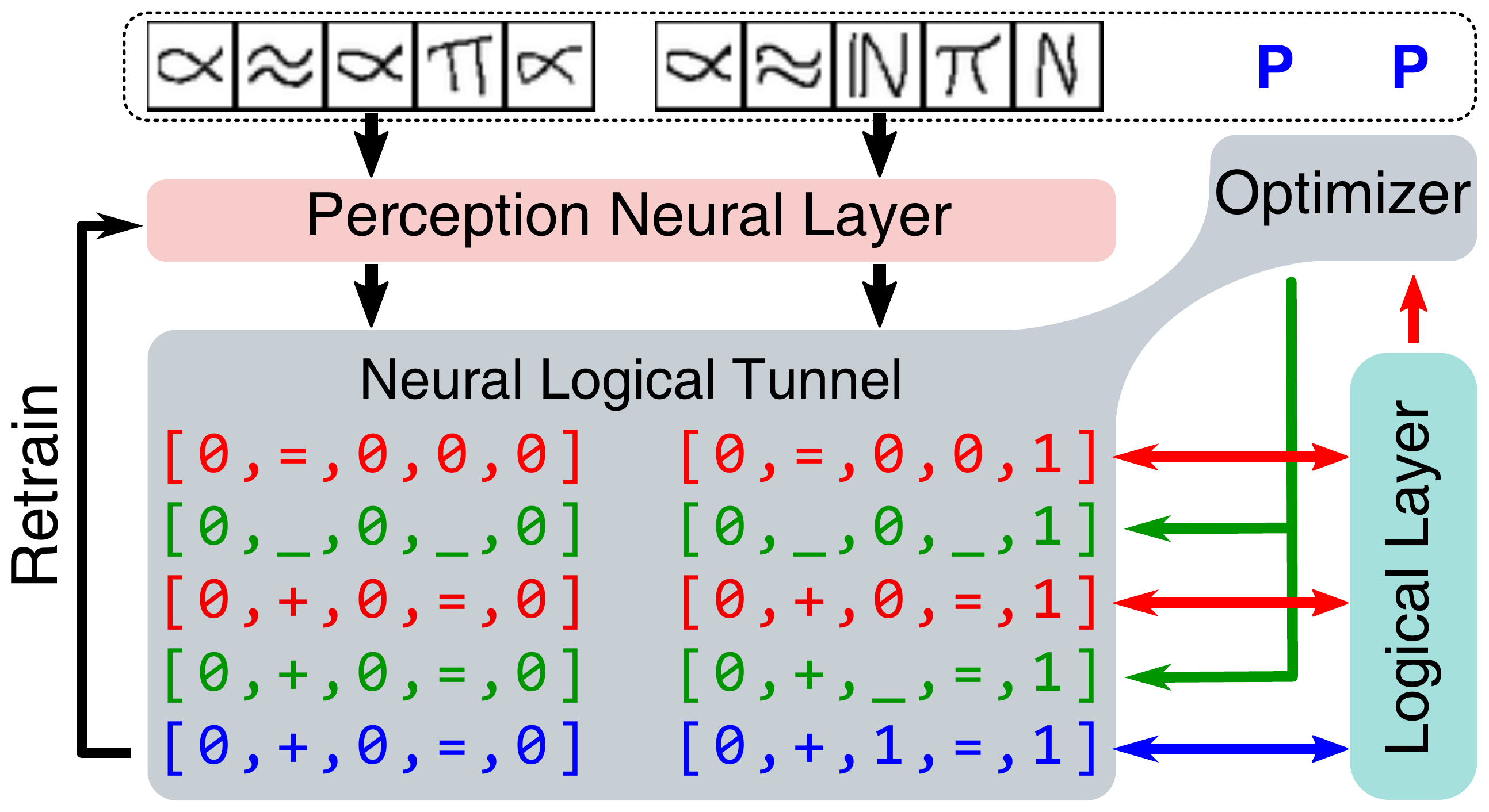}
    \caption{\textbf{Example of using logical abduction to correct the perception layer}. First, the perception neural layer incorrectly interprets the two images of positive examples and feeds them to the neural logical tunnel (the downward black arrows). Then, the logical layer finds them inconsistent (red arrows) and makes a request to the derivative-free optimization to substitute some digits into “blank variables” (green arrows). Finally, the logical layer successfully abduces a consistent interpretation for the two images (the blue arrow) and uses them as labels of the two images to retrain the perception neural layer (the upward black arrow).}
    \label{fig:percept}
\end{figure}

\textbf{Abductive logic programming.} Abduction refers to a reasoning process of forming a hypothesis that explains given observed phenomena according to domain knowledge~\cite{peirce:abduction}. For example, consider the following knowledge written as first-order logical formulas:
\begin{align}
    \texttt{wet\_grass}&\texttt{ :- }\texttt{rain\_last\_night}.\label{eq:grass1}\\
    \texttt{wet\_grass}&\texttt{ :- }\texttt{sprinkler\_was\_on}.\label{eq:grass2}\\
    \texttt{wet\_shoes}&\texttt{ :- }\texttt{wet\_grass}.\label{eq:shoe}\\
    \texttt{\bf false}&\texttt{ :- }\texttt{rain\_last\_night,sprinkler\_was\_on}.\label{eq:ic}
\end{align}
where the first three formulas state the causes for grass and shoes being wet, and the last formula specifies that the given two conditions cannot be true at the same time. When an observation of \texttt{wet\_shoes} is true, formula~\ref{eq:shoe} is regarded as an explanation, indicating \texttt{wet\_grass} should also be true. Continuing this process, both \texttt{rain\_last\_night} and \texttt{sprinkler\_was\_on} are other possible explanations. If it is observed that no rain occurred last night, according to the constraint in Formula~\ref{eq:ic}, \texttt{sprinkler\_was\_on} would be the only explanation.

A declarative framework in Logic Programming that formalizes this process is Abductive Logic Programming (ALP)~\cite{KakasKT92ALP}. In this framework, an abductive logic theory $T$ is a triple $(KB,A,IC)$, where $KB$ is a knowledge base of domain knowledge, $A$ is a set of abducible predicates or propositions, and $IC$ is the integrity constraints of the theory. The logic program $KB$ consists of a set of first-order logic formulas that describes the domain, including complete definitions for a set of observable predicates or propositions, and a set of abducible predicates or propositions $A$ that have no definitive rules in $T$. For the above example, $KB$ involves Formulas \ref{eq:grass1} to \ref{eq:shoe}, $A$ consists of the two propositions without any definitive formula: $\{\mathtt{rain\_last\_night}, \mathtt{sprinkler\_was\_on}\}$, and Formula~\ref{eq:ic}  is the integrity constraint. Formally, an abductive logic program can be defined as follows~\cite{KakasKT92ALP}:

\begin{definition} Given an abductive logic theory $T=(KB, A, IC)$, an abductive explanation for observed data $X$, is a set, $\Delta$, of ground abducibles of $A$, such that:
    \begin{itemize}
      \item $KB\cup\Delta\models X$
      \item $KB\cup\Delta\models IC$
      \item $KB\cup\Delta$ is consistent.
    \end{itemize}
    where $\models$ denotes the logical entailment relation.
\end{definition}
Intuitively, the abductive explanation $\Delta$ serves as a hypothesis that explains how an observation $X$ could hold according to the logical theory $T$.

In the binary additive learning tasks shown in Figure~\ref{fig:data}, the abductive theory $T$ contains a set of first-order logical rules for parsing symbol lists into equations and performing bitwise calculations, as shown in Figure~\ref{fig:BK}. Specifically, the components are listed as follows.

The domain knowledge $KB$ first contains rules for parsing a list of symbols to an equation. By assuming that all the equations in the data have the form \texttt{X+Y=Z}, this piece of domain knowledge can be expressed with a Prolog DCG formula:
\begin{align}
    \texttt{eq --> digits,[+],digits,[=],digits.}\label{eq:parser}
\end{align}
where \texttt{eq} is a list of symbols such as \texttt{[1,+,1,0,1,=,1,1,0]}, \texttt{digits} represents a list of digital symbols, for example \texttt{[0]} and \texttt{[1,0,0,1]}. This parser can parse the list \texttt{eq} into a Prolog term \texttt{calc(X,Y,Z)}, where the variables correspond to the parsed \texttt{digits}.

To enable arithmetic calculation in logical layer, the domain knowledge $KB$ also include certain rules for calculating a parsed equation, e.g., \texttt{calc(X,Y,Z)}. We implemented it using a bitwise additive calculator with the following Prolog formula:
\begin{align}
    \texttt{calc(X,Y,Z) :- op\_rules(R),bitwise\_calc(R,X,Y,Z).}\label{eq:calc}
\end{align}
where \texttt{op\_rules} is a predicate that declares a list of \emph{unknown}  bitwise operations as addition rules that can be applied to \texttt{bitwise\_calc} to perform calculations.

One type of abducibles is the bitwise operations described with the predicate \texttt{my\_op(D1,D2, [D3])}, which represents a bitwise addition rule \texttt{D1+D2=D3}. For example, a logical exclusive-or operation can be defined with a list of bitwise operations\texttt{[my\_op(0,0,[0]),my\_op(1,0,[1]),} \texttt{my\_op(0,1,1),my\_op(1,1,[0])]}, and the carry rule for arithmetic addition can be written as \texttt{my\_op(1,1,[1,0])}. Note that these bitwise addition rules are not included in the domain knowledge---they will be abduced as explanatory hypotheses for the training data during learning process.

Another type of abducibles involves the lists of symbols \texttt{eq} in Formula~\ref{eq:parser}, which are the input to the logic layer through the neural-logical tunnel. Typically, the original \texttt{eq$_0$=[l$_1$,$\ldots$,l$_s$]} interpreted by initial perception model would contain mistakes and cause failures when attempting to abduce consistent hypotheses. Therefore, the neural-logical tunnel will try to substitute some \texttt{l$_i$} with a Prolog variable ``\texttt{\_}'' as blanks in the equation. ALP will then abduce symbols that satisfy the consistency constraints to fill in these blanks. For example, when the perception model is under-trained, the neural-logical tunnel is highly likely to receive an \texttt{eq$_0$=[1,1,1,1,1]}, i.e., the perception model interprets the image of the equation as ``11111'', which is definitely inconsistent with any arithmetic rules. Observing that ALP cannot abduce a consistent hypothesis, the neural logical tunnel will begin substituting some of the values in \texttt{eq$_0$} with blank Prolog variables, e.g., \texttt{eq$_1$=[1,\_,1,\_,1]}. Then, ALP can abduce a consistent hypothesis involving the additive rule\texttt{my\_op(1,1,[1])} and \texttt{eq$'_1$=[1,+,1,=,1]}. Finally, the abduced \texttt{eq$'_1$} can be used as a supervised signal to train the perception model, helping it distinguish images of ``+'' and ``='' from other symbols. An example of this process is illustrated in Figure~\ref{fig:percept}.

The abduced answer $\Delta$ contains hypotheses of the previous two abducibles (i.e., a list of \texttt{my\_op} rules for the \texttt{op\_rules} predicate in Formula~\ref{eq:calc} as the reasoning model $\Delta_{C}$, and a list of (modified) digit and operator symbols \texttt{eq$'$} for retraining the perception model).

The integrity constraint $IC$ simply addresses the consistency of the abduced hypotheses. For example, the bitwise operations  \texttt{my\_op(1,0,[1])} and \texttt{my\_op(1,0,[0])} cannot be valid bitwise addition rules at the same time; \texttt{my\_op(1,0,[0])} and \texttt{eq$'$=[1,+,0,=,1]} cannot be both be abduced output as an explanatory hypothesis.

The observed fact $X$ is the entire training dataset, which consists of all the images of equations and the labels indicating their correctness. However, in first-order logic, evaluating the consistency of a set of formulas on given facts is NP-hard. Hence, during the abduction process it will be difficult to evaluate the consistency of an abduced hypothesis on the entire training set. Therefore, NLM performs abduction on subsampled data over multiple iterations, where each subsample contains only 5---10 equation images. The abduced locally consistent hypotheses in each subsample are saved as relational features.

\textbf{Optimization}. The learning target of NLM’s abduction is to find a hypothesis $\Delta$ that maximizes its consistency $Con(\Delta,X)$ on a set of observed examples $X=\{x_1,\ldots,x_n\}$. This objective can be written as follows:
\begin{align}
    \argmax\limits_{\Delta}\quad Con(\Delta,X)\label{eq:obj_abdc}
\end{align}
where the consistency $Con(\Delta,X)$ is defined by the size of the maximum consistent subset $X_C\subseteq X$ derived from $\Delta$, which can be formalized as follows:
\begin{align}
    \argmax\limits_{X_c\subseteq X}\quad&|X_c|\label{eq:obj_con}\\
    \mathrm{s.t.}\quad&KB\wedge\Delta\models X_c.\nonumber\\
    &P\wedge\Delta\models IC.\nonumber
\end{align}
where $KB$ and $IC$ are the domain knowledge and integrity constraints, respectively, defined in the abductive logic program of the logical layer. The optimization problem in objective~\ref{eq:obj_con} is a subset-selection problem, which is also generally NP-hard. Therefore, we approximately solve objectives~\ref{eq:obj_abdc} an~\ref{eq:obj_con} with greedy algorithms.

As shown in Figure~\ref{fig:percept}, when the perception model is under-trained, the perceived symbols \texttt{eq$_0$} from $X$ might contain mistakes, causing the abduction of sufficiently consistent hypotheses to fail. NLM tries to solve this problem by substituting some possibly incorrectly perceived symbols in \texttt{eq$_0$} to blank variable ``\texttt{\_}'' and lets ALP abduce a symbol list \texttt{eq$'_1$} that ensures a maximally consistent $\Delta$ on dataset $X$; then, it retrains the perception model. The substitution vector can be represented by $S=\{0,1\}^l$, where $l$ is the length of the interpreted symbol list \texttt{eq$_0$} from $X$. When $S_i=1$ then the $i$-th interpreted symbol in \texttt{eq$_0$} will be replaced with a blank variable ``\texttt{\_}''. When abducing hypotheses that are too far away from the perceived symbols and obtaining trivial solutions, the number of substituted variables should be constrained. Thus, the objective can be formalized as follows:
\begin{align}
    \max_{S\in\{0,1\}^l}\max_{\Delta(S)}\quad&Con(\Delta(S),X)\label{eq:obj_subs}\\
    \mathrm{s.t.}\quad&\left \| S \right \|_0 \leq k, k>1.\nonumber
\end{align}
where $\Delta(S)$ is a hypothesis abduced by ALP whose symbolic interpretations \texttt{eq$_0$} are modified with the substitution vector $S$, and $k$ is the limit on the number of modified perceived symbols. In the experiments we set $k=2$. 

The optimization problem in~\ref{eq:obj_subs} is a binary vector optimization problem in an extremely complex hypothesis space, popular gradient-based optimization techniques can hardly be applied to this scenario. Therefore, we adopt a derivative-free optimization technique, RACOS~\cite{Yu2016Derivative}, to solve it. RACOS is a randomized derivative-free optimization method implemented by a classification model that discriminates good solutions from bad ones, and it achieves good performance on complex optimization problems.

\textbf{Making decisions}. The high complexity of the optimization objective in Equation~\ref{eq:obj_subs} makes it infeasible for the NLM to evaluate the entire training set $X$ during optimization. Therefore, NLM performs abduction and optimization for $T$ times, using a small observed dataset $X_t\subseteq X$ subsampled from the original dataset $X$ each time. While the perception model $p$ is iteratively trained from $t=1$ to $T$, the locally consistent reasoning model $\Delta_C^t$ in each iteration cannot be simply replaced or merged to construct the final output reasoning model $\Delta_C$. Because the training data in each iteration $X_t\neq X$, there is no guarantee that $\Delta_C^t$ will be consistent for all training examples $x\in X$. In fact, some sets of examples (such as the arithmetic equations ``1+10=11, 11+100=111'') can achieve maximum consistency using just one bitwise addition rule \texttt{my\_op(1,0,[1])}, which results in an incomplete $\Delta_C^t$ for defining the target concept. Moreover, when the perception model $p_t$ in the $t$-th turn is under-trained, ALP might abduce incorrect bitwise addition rules.

To solve this problem, inspired by the propositionalization technique in Inductive Logic Programming~\cite{kramer:propositionalisation}, we retain the reasoning models abduced in each iteration as a \emph{relational feature}. Before the perception model is well-trained, NLM uses a buffer that retains only the latest $R$-learned $\Delta_C^t$, and---based on their performances---some of these will be discarded during the training of the decision neural layer.

When an equation $x_i$ is input into NLM, its symbolic interpretation \texttt{eq$^t$} mapped by the perception model will be evaluated by all the relational features to produce a binary vector $r_i=[r_{i1},\ldots,r_{iR}]$, where
\begin{equation}
    r_{ij}=\left\{
    \begin{aligned}
      1,\quad&KB\wedge\Delta_C^j\models x_i,\\
      0,\quad&KB\wedge\Delta_C^j\not\models x_i.
    \end{aligned}
    \right.
\end{equation}
This vector of relational features transforms the original dataset $X=\{\langle x_i,y_i\rangle\}$ into a new dataset $X'=\{\langle r_i,y_i\rangle\}$, from which a decision model is learned by the decision layer in Figure~\ref{fig:architecture}.

On one hand, the retained relational features are still first-order logical rules serving as a reasoning model with good human comprehensibility. On the other hand, using the propositionalization technique in the decision learning process reduces the impact of the noise introduced by the random subsampling of the training data.


\subsection*{Experiment: Handwritten Equations Decipherment}
\begin{figure}
    \centering
    \includegraphics[width=0.8\linewidth]{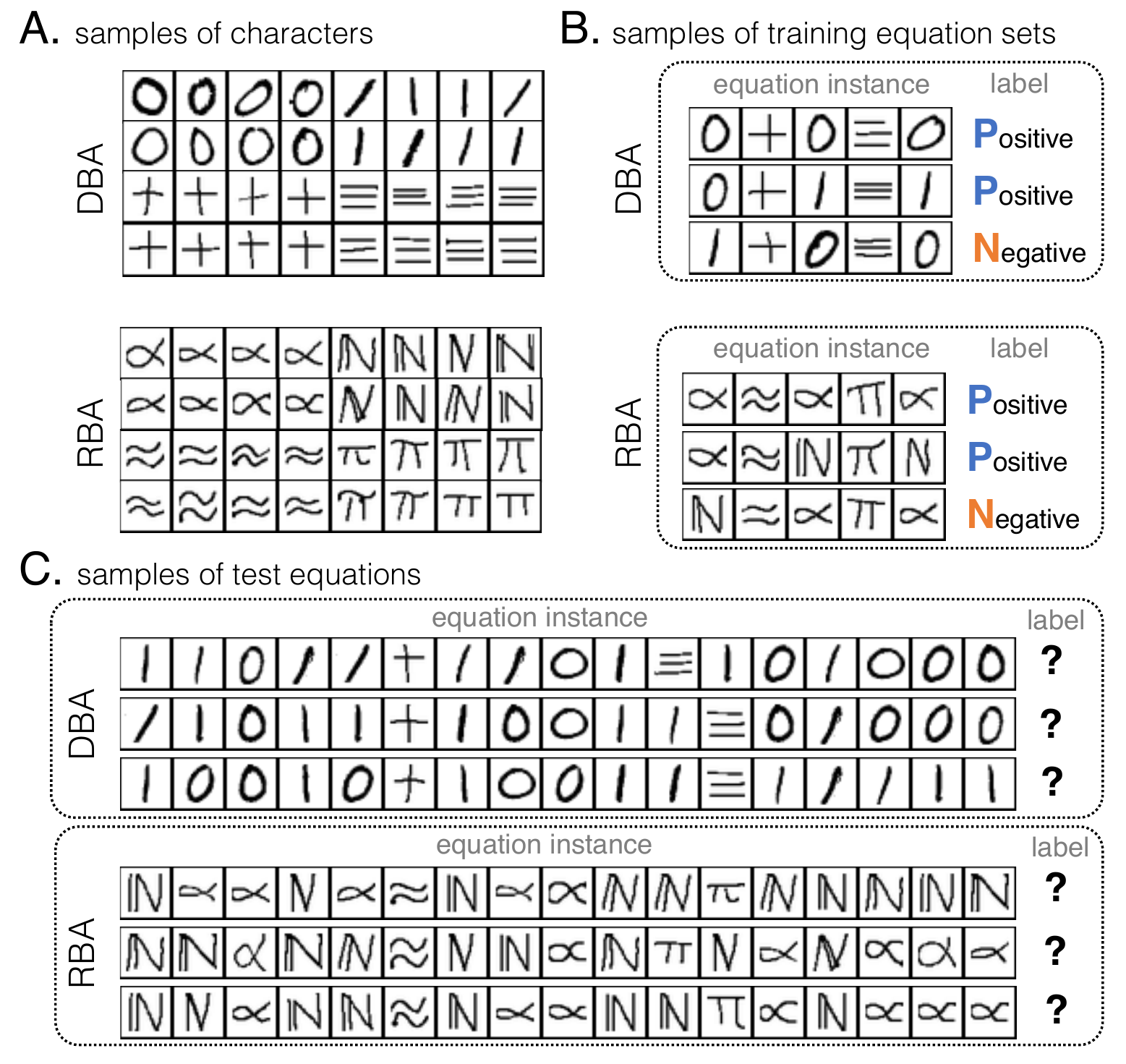}
    \caption{\textbf{Dataset illustrations}. Our datasets are constructed from images of symbols (``0'', ``1'', ``+'' and ``=''). \textbf{(A)} Samples of the two types of characters corresponding to those symbols: The Digital Binary Additive (DBA) set and the Random Binary Additive (RBA) set. \textbf{(B)} Samples of training equation sets of length 5, where each equation is associated with a label: \emph{positive} or \emph{negative}. A learning algorithm is required to learn from a set of equations and their labels and then predict the labels of testing equations with different lengths. \textbf{(C)} Samples of test equations of length 17, where labels are not shown to the algorithms; the labels are used only to calculate the accuracy of the predictions.}
    \label{fig:data}
\end{figure}

\begin{figure}
    \centering
    \includegraphics[width=.7\linewidth]{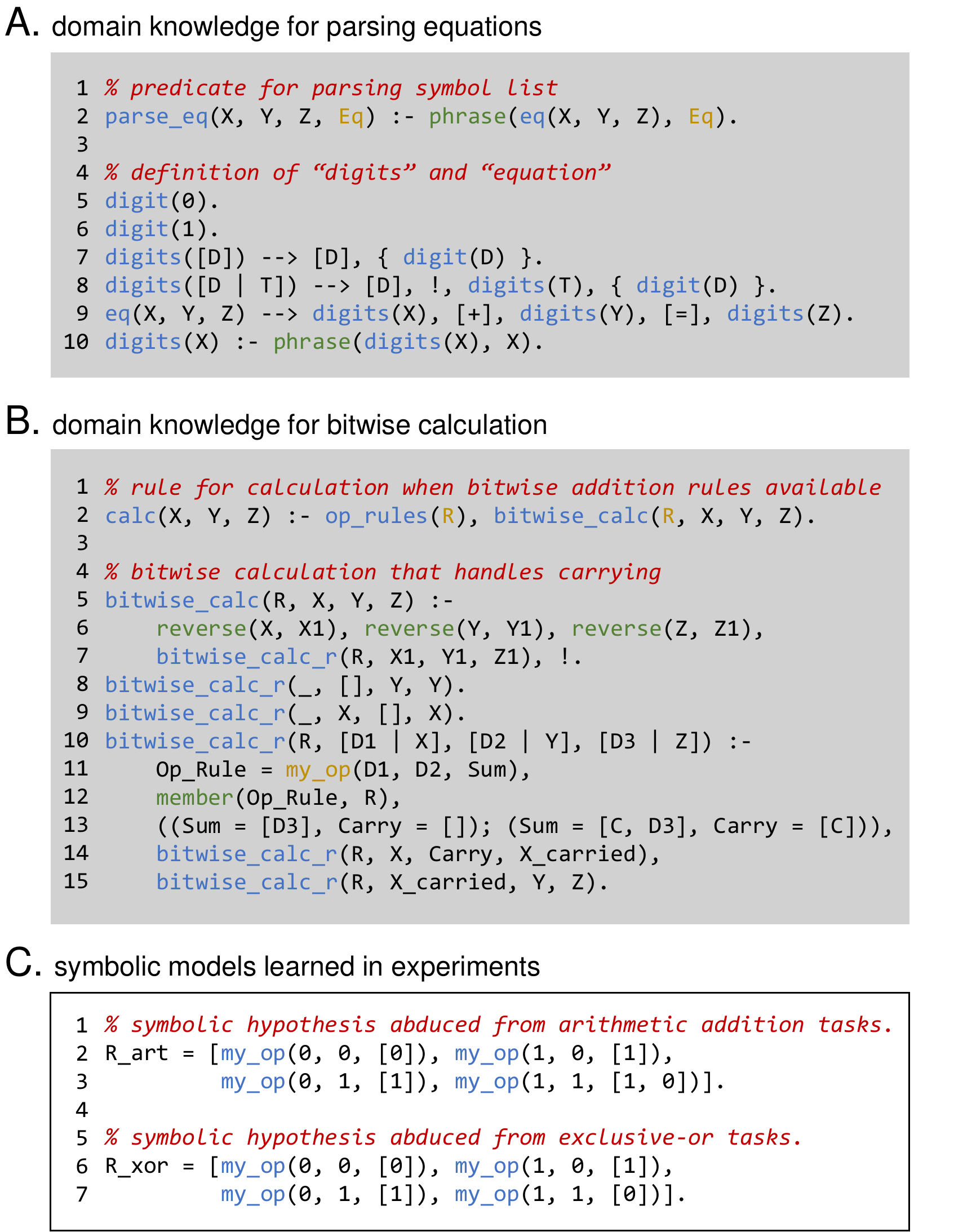}
    \caption{\textbf{Domain knowledge as Prolog programs used in the logical layer}. The blue and green words are user-defined and system predicates, respectively, while the yellow words are abducibles that can be derived from the observed data. \textbf{(A)} A Prolog program that can parse and abduce a list of symbols as equations. For example, when  \texttt{Eq=[1,\_,1,\_,1]} (the ``\texttt{\_}'' is a Prolog ``blank'' variable that can be filled with an abduced symbol), using ALP ``\texttt{parse\_eq}'' can abduce ``\texttt{X=Y=Z=1}'' and ``\texttt{Eq=[1,+,1,=,1]}''. \textbf{(B)} A Prolog program for making bitwise calculations and abducing bitwise operations as learned reasoning models. When a positive example perceived as ``\texttt{[1,1,+,1,=,1,0,0]}'' is provided, using ALP it can abduce two bitwise addition rules ''\texttt{op\_rules([my\_op(1,1,[1,0]),my\_op(1,0,[1])])}''). \textbf{(C)} Examples of output reasoning models (relational features) abduced by NLM from data; different hypotheses can be learned from different tasks.}
    \label{fig:BK}
\end{figure}

We constructed two image sets of symbols to build the equations shown in Figure~\ref{fig:data}. The Digital Binary Additive (DBA) equations were created with images from benchmark handwritten character datasets~\cite{mnistdata,symboldata}, while the Random Binary Additive (RBA) equations were constructed from randomly selected character sets and shared isomorphic structure with the equations in the DBA tasks. Each equation is input as a sequence of raw images of digits and operators. As an analogy to the Mayan hieroglyph decipherment by historians, we provided NLM with background domain knowledge of arithmetic structure rules as shown in Figure~\ref{fig:BK}(A) and~\ref{fig:BK}(B). Note that this knowledge does not specify the type of calculation in the equations; instead, NLM needs to learn that from the data. Examples appear in Figure~\ref{fig:BK}.

In the NLM implemented for this task, the perception layers consist of a two-layer CNN and a two-layer multiple-layer perceptron (MLP) followed by a softmax layer; the logical layer will abduce 20 bitwise operations set as relational features; The decision neural layer is a two-layer MLP. We test the learning performance of NLM by comparing it with the Bidirectional Long Short-Term Memory network (BiLSTM)~\cite{hochreiter1997long} and the Differentiable Neural Computer (DNC)~\cite{Graves2016Hybrid}, because both are state-of-the-art benchmark models capable of solving tasks from sequential input such as arithmetic equations. In particular, the DNC has shown its potential on symbolic computing tasks~\cite{Graves2016Hybrid} because it is associated with memory. To handle the image inputs, BiLSTM and DNC also use a CNN as their input layers. The training data prepared for BiLSTM and DNC contains equations with lengths from 5 to 26, but for NLM, we used only equations with lengths from 5 to 8, where each length contains 300 random sampled equations. In the testing stage, all the methods are tasked with predicting 6,600 equations whose lengths range from 5 to 26, where each length contains 300 examples.

\begin{figure}
    \centering
    \includegraphics[width=1\linewidth]{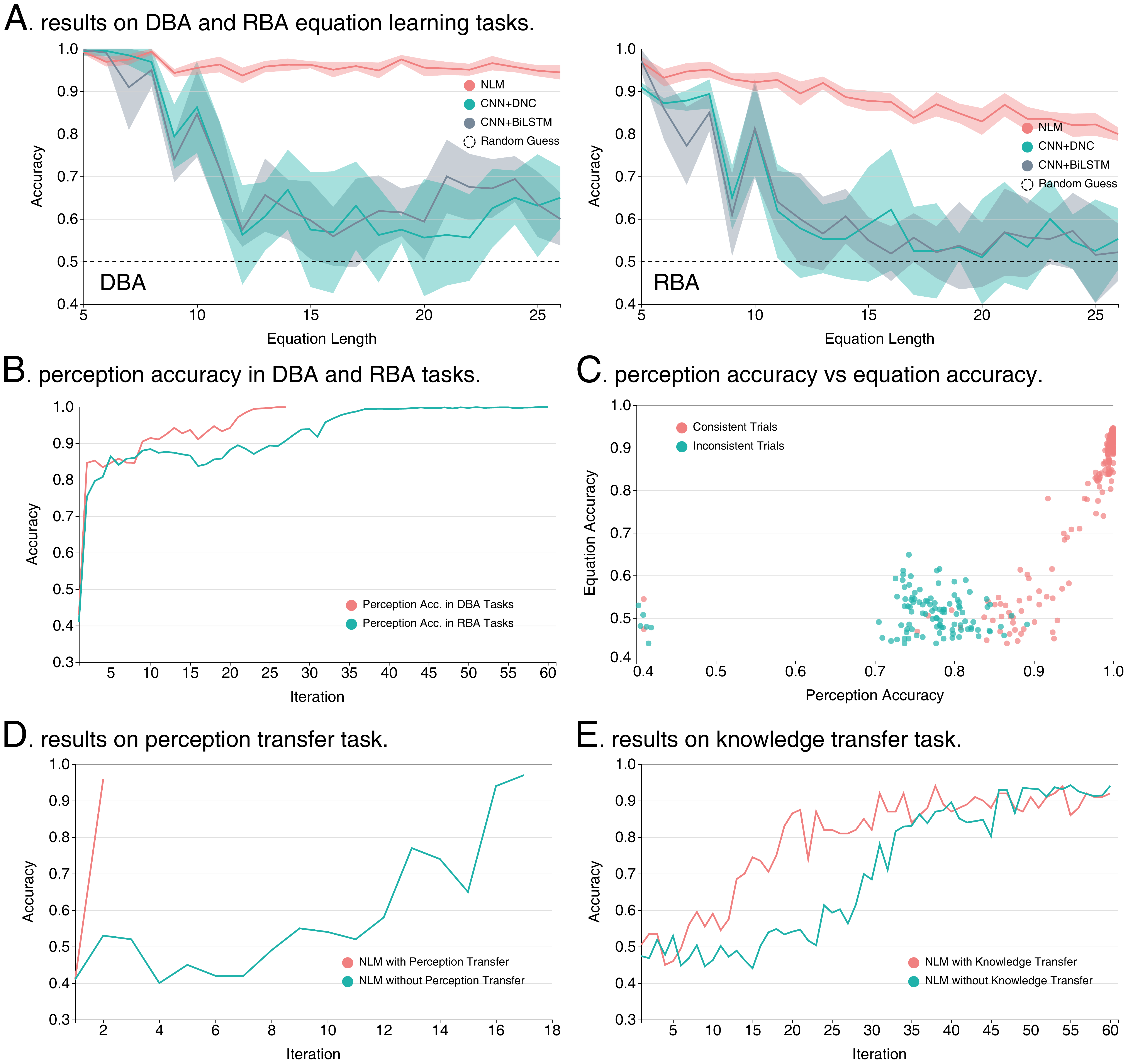}
    \caption{\textbf{Experimental results}: \textbf{(A)} Performances of the NLM and the compared methods on the DBA and RBA equation classification tasks, respectively. The values on the x-axis show the lengths of the test equations; the shadowed areas represent standard deviations. \textbf{(B)} The average training accuracy of the perception neural network vs. the number of iterations on the DBA and RBA tasks. \textbf{(C)} The relationship between the accuracy of character perception and equation classification during training. Each dot represents an abductive problem-solving trial with subsampled data: the red dots are trials where the NLM successfully found consistent symbolic hypotheses. \textbf{(D)} and \textbf{(E)} Performances of the NLMs on the perception and knowledge transfer tasks.}
    \label{fig:results}
\end{figure}

Figure~\ref{fig:results}(A) shows that on both tasks, the NLMs significantly outperform the compared methods while using far fewer training equations with lengths of at most 8. All the methods performed better on the DBA task than on the RBA task, because the symbol images in the DBA task are more easily distinguished. The performance of BiLSTM and DNC degenerates quickly toward the random-guess line as the length of the testing equations grows, while the performance of NLM degenerates much more slowly and maintains an accuracy above 80\%. These results verify that NLM has learned the correct rules from the data. The logical layer benefits considerably from its ability to handle first-order logic; thus, it naturally exploits the definitions of the symbolic primitives and the background knowledge when solving the task. Note that the background knowledge exploited by NLM in this task involves no more than the logic structure rules, which alone cannot define the target concept; however, only the symbolic rules must be learned from data. The learned rules are shown in Figure~\ref{fig:BK}(C). As a result, NLM generalizes well from the training data distribution, exhibiting a learning ability closer to that of humans. 

Inside the learning process of NLM, although no labels exist for the images of digits and operators, the perception accuracy did increase during the learning process, as shown by Figure~\ref{fig:results}(B). The results verified that logic consistency can be very useful for providing a surrogate supervised signal and that, through logical abduction, NLM can correct incorrect neural perceptions. Figure~\ref{fig:results}(C) shows the relationship between the overall equation classification accuracy and the image perception accuracy during NLM training on the RBA tasks, where each dot in the figure represents a trial of abducing a consistent hypothesis from a subsample of equations; the red dots indicate successful abductions and the green dots signify failures. From the results, it is apparent that a better perception accuracy indeed contributed to better equation classification accuracy.

\subsection*{Experiment: Cross-task Transfer}
Just as when Bowditch used his past knowledge about Mayan hieroglyphs and calendar units to decipher unknown scripts, human learning features the ability of cross-domain transfer. We further investigate whether NLM, with its improved learning ability, also has an improved transfer capability.
The first task investigates the transferability of the perception model to domains with different symbolic structures. We created logical exclusive-or equations as training examples using the DBA characters; the size of the dataset was equal to that used for the binary additive equation learning tasks.

To transfer the perception model to the exclusive-or domain, the perception layers used in the NLM are a direct copy of those trained on the DBA task. During the learning process, the parameters of those layers are fixed, ensuring that the NLM learns only reasoning models. As a comparison, another NLM was trained on the exclusive-or task from scratch. The results are shown in Figure~\ref{fig:results}(D), and the learned reasoning model is shown in Figure~\ref{fig:BK}(C). As we can observe from the result, the final performances of NLMs with and without perception transfer are comparable. However, the convergence of the NLM with perception transfer is much faster than that without the transfer. This is consistent with the results in Figure~\ref{fig:results}(C) and shows that a good perception model dramatically reduces the difficulty of learning reasoning models.

The second task attempted to evaluate the capability of NLMs for transferring learned symbolic knowledge. We did this by setting the source and target domain as the RBA and DBA tasks, respectively. The knowledge transferred is in the form of both the logical layer and the decision neural layer. The NLM with knowledge transfer copied the relational features and the decision MLP parameters from a well-trained RBA NLM, therefore it only needed to train its perception CNN, in which the list of symbols abduced by the logical layer are regarded as labels to the input images. Similar to the previous experiment, the compared NLM learns the entire model from scratch. From the results depicted in Figure~\ref{fig:results}(E), we can observe that the NLMs eventually reach the same level of accuracy; however, the NLM with knowledge transfer converged significantly faster than the compared method. This result verifies that well-built domain knowledge makes learning easier. However, comparing the results between knowledge transfer and perception transfer, we can see that providing the learning perception model without explicitly providing the labels is considerably more difficult---which indeed caused considerable trouble when historians were trying to decipher the Mayan language.

\subsection*{Conclusion and Discussion}
The experimental results verified that \emph{abductive learning} can perform human-like abductive problem-solving that combines neural perception and symbolic reasoning. Abductive learning works even when the given symbolic background knowledge is incomplete for learning the target concepts. The proposed NLM approach can exploit symbolic domain knowledge while processing sub-symbolic data such as raw pixel images. In our experiments, the only supervision exploited by NLM involves labels of high-level concepts such as the ``correctness of an equation''. For mid-level concepts such as the digits and operator symbols that serve as primitives for high-level reasoning, NLM learns a recognition model without requiring image labels.

To the best of the authors' knowledge, abductive learning is the first framework designed for simultaneously learning both reasoning and perception models. To accomplish this goal, the AI system must be able to simultaneously manipulate symbolic learning and sub-symbolic learning. In past AI research, these two abilities have been developed only separately.

Symbolic AI has been considered as a fundamental solution to artificial intelligence since the dawn of AI. Symbolic AI refers to a set of methods based on high-level ``symbolic'' problem representations and its goal is to define intelligent systems in an explicit way that is understandable by humans. In this branch of AI research, a series of important steps has been achieved, e.g., automatic theorem proving~\cite{NewellS56,Chang:1997:SLM}, propositional rule learning~\cite{FurnkranzGL12}, expert systems~\cite{Jackson99ES}, automated planning~\cite{FikesN71STRIP} and inductive logic programming~\cite{mugg:ilp}, and so on.

By formalizing the problem-solving process using symbolic representations such as a first-order logic language, symbolic AI imitates human reasoning through symbolic computation---based mainly on heuristic and selective search. Owing to the computer’s high efficiency in solving searching problems, symbolic AIs can deal with many tasks that are difficult for humans~\cite{simon1971human}. For example, recent progress in playing the game of Go verified that learning algorithms taught by heuristically searched examples perform even better than those trained from demonstrations involving human expertise~\cite{alphagozero}. In fact, the success of symbolic AI over the last century has already shown the advantages of using computers to solve many symbolic reasoning tasks as compared to humans. 

Most of the current mainstream machine learning methods, such as statistical learning and deep neural networks, focus on problems that have continuous and sub-symbolic hypothesis space representations to which gradient-based optimization techniques can be easily applied. Consequently, these methods have achieved great success on perception-like tasks such as image recognition~\cite{Krizhevsky2012ImageNet,NIPS2013_5207}, speech recognition~\cite{Hinton2012Deep} and so on. However, due to the limited expressive power of sub-symbolic representation, most of these machine learning methods are incapable of performing secondary reasoning on learned data. For example, although deep neural networks have achieved near-human performance on some simple natural language processing tasks, a recent empirical study noted that they cannot address reading comprehension problems~\cite{Jia2017EMNLP}. Moreover, the gap between symbolic and sub-symbolic representations causes them to fail to make use of domain knowledge that can only be expressed through first-order logic; thus, they miss an opportunity for exploiting the great inventions in the fruitful symbolic AI research. 

A natural approach to fix these problems is to combine symbolic and sub-symbolic AI systems. Fuzzy logic~\cite{Zadeh65:fuzzy}, probabilistic logic programming~\cite{deraedt:pilp} and statistical relational learning~\cite{getoor:srlbook} have been proposed to empower traditional logic-based methods to handle probability distributions; however, most of them still require human-defined symbols as input similar to traditional symbolic AI systems. Probabilistic programming is then proposed as an analogy to human cognition to enable probabilistic reasoning with sub-symbolic primitives~\cite{Tenenbaum:probprog,Lake:humanlevel,kulkarni:picture}, yet the correspondence between the sub-symbolic primitives and their symbolic representations used in programming is assumed to already exist rather than assuming that it should be automatically learned~\cite{Lake:humanlevel,kulkarni:picture}. Differential programming methods such as DNC attempt to emulate symbolic computing using differentiable functional calculations~\cite{Graves2016Hybrid,GauntBKT17,BosnjakRNR17}, but they can hardly reproduce genuine logical inferences, and they require large amounts of training data to learn even simple logical operations. LASIN is a work that also uses logical abduction to introduce general human knowledge into statistical learning~\cite{daiLASIN}; however, the exploited symbolic knowledge is required to be complete and correct. In short, few AI systems exist that can perform symbolic and sub-symbolic learning at the same time. 

As shown in the Mayan digits decipherment example, humans can easily combine knowledge-based symbolic reasoning and sensory-based sub-symbolic perception, and these two components affect each other simultaneously~\cite{solso2008cognitive}. This fact has been verified in cognitive psychology. The most representative evidence is visual illusions, which illustrate that the way we understand the world is greatly influenced by the complicated interactions between our sensations and our past knowledge that lend abundant meaning to sensed stimuli~\cite{solso2008cognitive}. It has been suggested that abduction is the key to the entangled relationship between sub-symbolic perception and symbolic reasoning connections~\cite{Magnani:2009:abd}. In a sense, we believe that the proposed abductive learning can be regarded as a verification of this theory. 

Inspired by the trial-and-error approach of Bowditch's decipherment (Figure~\ref{fig:puzzle}(B)), abductive learning connects symbolic reasoning with sub-symbolic perception. Owing to the expressive power of first-order logic, abductive learning is capable of directly exploiting general domain knowledge. The differentiable structure of neural perception enables abductive learning to conveniently create the mappings between raw data and its symbolic representations. Finally, the background knowledge based heuristic search allows abductive learning to automatically infer the existence of primitive objects from raw data. 

Many improvements can be made to abductive learning in the future. Ideally, an AI system should be able to learn the background knowledge by itself; in addition, the learned knowledge should be reusable in other tasks. In this work, we verified that both perception model and reasoning model can be directly reused in different but similar tasks; however, a general paradigm for reusing more complicated knowledge is required. Moreover, loss of the learning system is currently back-propagated through its logical module by searching, abduction and optimization in a naive way. Methods for improving the efficiency of these operations are worth studying and can make abductive learning more practical for real applications. 


\bibliography{ourbib}
\bibliographystyle{acm}

\end{document}